\documentclass[11pt,a4paper]{article}
\usepackage[hyperref]{acl2020}
\usepackage{times}
\usepackage{graphicx}
\usepackage{latexsym}

\usepackage{microtype}

\aclfinalcopy % Uncomment this line for the final submission
\title{What BERT Sees: Cross-Modal Transfer for \\ Visual Question Generation}

\author{Thomas Scialom$^{\star\ddagger}$\thanks{$^{\ast}$: equal contribution.} \quad Patrick Bordes$^{\ddagger\ast}$ \quad Paul-Alexis Dray$^{\star}$ \quad \\{\bf Jacopo Staiano$^{\star}$ \quad Patrick Gallinari$^{\ddagger}$} \\
$^\ddagger$ Sorbonne Universit\'e, CNRS, LIP6, F-75005 Paris, France\\
$^\star$ reciTAL, Paris, France \\
  {\tt \{thomas, paul-alexis, jacopo\}@recital.ai} \\
  {\tt\{patrick.bordes, patrick.Gallinari\}@lip6.fr}
  \\}

\date{}

\begin{document}
\maketitle
\begin{abstract} 

Pre-trained language models have recently contributed to significant advances in NLP tasks. Recently, multi-modal versions of BERT have been developed, using heavy pre-training relying on vast corpora of aligned textual and image data, primarily applied to classification tasks such as VQA. In this paper, we are interested in evaluating the visual capabilities of BERT \textit{out-of-the-box}, by avoiding pre-training made on supplementary data. We choose to study Visual Question Generation, a task of great interest for grounded dialog, that enables to study the impact of each modality (as input can be visual and/or textual). Moreover, the generation aspect of the task requires an adaptation since BERT is primarily designed as an encoder.  We introduce \emph{BERT-gen}, a BERT-based architecture for text generation, able to leverage on either mono- or multi- modal representations. The results reported under different configurations indicate an innate capacity for \emph{BERT-gen} to adapt to multi-modal data and text generation, even with few data available, avoiding expensive pre-training. The proposed model obtains substantial improvements over the state-of-the-art on two established VQG datasets.

\end{abstract}

\section{Introduction}

In Artificial Intelligence, several works have investigated the longstanding research question of whether textual representations encode some sort of visual information. 
This has been done primarily for word embeddings, e.g, by applying them to Zero-Shot Learning \cite{DBLP:conf/icml/ZablockiBSPG19}, or sentence embeddings, e.g, by applying them to Image Captioning \cite{DBLP:journals/tacl/SocherKLMN14}.
In this paper, we are interested in evaluating the visual capacities of \textit{pre-trained language models}; in our case, BERT \cite{devlin2019bert}. 

To do so, we choose the Visual Question Generation (VQG) \cite{DBLP:conf/acl/MostafazadehMDM16} task, for the following reasons.
First, from a practical standpoint, the VQG task has several applications: robots or AI assistants could ask questions rooted in multi-modal data (e.g. fusing conversational data with visual information from captors and cameras), in order to refine their interpretation of the situation they are presented with. 
Second, it could also allow systems relying on knowledge-bases to gain visual common sense and deal with the Human Reporting Bias \cite{DBLP:conf/cvpr/MisraZMG16}, which states that the content of images and text are intrinsically different, since visual common sense is rarely explicitly stated in text. 
Moreover, unlike Image Captioning (where the input is only visual) or VQA (where the input is visual \textit{and} textual), VQG is a multi-modal task where input can be textual \textit{and/or} visual: this is of particular interest to analyze the impact of each modality. 
Finally, VQG relies on textual generation, which is challenging since BERT is not primarily designed for generation.

BERT-based Multi-Modal Language Models have been proposed 
\cite{DBLP:journals/corr/abs-1908-03557,DBLP:journals/corr/abs-1908-08530} 
to tackle multi-modal tasks, relying on heavy pre-training and large corpora of aligned textual and visual data. From these works, it is left to explore whether the cross-modal capacities come from the pre-traning, or are to some extent already encoded in BERT's representations.

It has recently been shown that BERT can generalize to another \textit{language}, with great results, in a zero-shot manner \cite{DBLP:journals/corr/abs-1910-11856}, i.e. without supervision between languages. 
In preliminary experiments, we extended this work to another \textit{modality}: we found out that, in VQG, without any supervision between the images and the questions, the cross-modal alignment was not successfully learnt. This discrepancy between multi-\textit{lingual} and multi-\textit{modal} results might find its root cause in the intrinsic semantic difference between textual and visual modalities \cite{DBLP:journals/jair/BruniTB14}.
Nonetheless, we hypothesize that BERT contains some abstractions that generalize across modalities. If so, it may transfer knowledge to the visual modality even with few training data rather than expensive pre-training and complex architectures.

Thus, in contrast with latter Multi-Modal BERT approaches, \textit{we explicitly avoid} using the following complex mechanisms:
(1) \textit{Multi-modal supervision}: we do not exploit explicit supervision between images and captions through a pre-training step;  
(2) \textit{Image-specific losses}: specific pre-training losses such as Masked RoI Classification with Linguistic Clues \cite{DBLP:journals/corr/abs-1908-08530} or
    sentence-image prediction \cite{DBLP:journals/corr/abs-1908-03557}; 
(3) \textit{Non-linearities}: 
    we explore a scenario in which the only learnable parameters, for aligning image representations to BERT, are those of a simple linear projection layer. 

Furthermore, to the best of our knowledge, this paper is the first attempt to investigate multi-modal text \textit{generation} using pre-trained language models. We introduce \textit{BERT-gen}, a text generator based on BERT, that can be applied both in mono and multi-modal settings. We treat images similarly to text: while a sentence is seen as a sequence of (sub)word tokens, an image is seen as a sequence of objects associated to their corresponding positions (bounding boxes). We show how a simple linear mapping, projecting visual embeddings into the first layer, is enough to ground BERT in the visual realm: text and image object representations are found to be effectively aligned, and the attention over words transfers to attention over the relevant objects in the image. 

Our contributions can be summarized as follows: (1)  we introduce \emph{BERT-gen}, a novel method for generating text using BERT, that can be applied in both mono and multi-modal settings;
(2) we report state-of-the art results on the VQG task;
(3) we show that the semantic abstractions encoded in pre-trained BERT can generalize to another modality without pre-training of any sort; 
(4) we provide extensive ablations and qualitative analyses to interpret the behavior of \emph{BERT-gen} under different configurations (mono- or multi- modal).  

\section{Related Work}

\paragraph{Multi-modal Language Models}
Following the successful application of BERT \cite{devlin2019bert}, and its derivatives, across a great majority of NLP tasks, several research efforts have focused on the design of multi-modal versions of BERT.
The first attempt was VideoBERT \cite{DBLP:journals/corr/abs-1904-01766}, a joint \textit{video} and text model pre-trained on a huge corpus of YouTube videos, where the video is treated as a ``visual sentence" (each frame being a ``visual word") processed by the BERT Transformer. 

Concerning models jointly treating information from images and text, visual features extracted from the image are used as ``visual words", and a \texttt{[SEP]} special token is employed to separate textual and visual tokens. In the literature, visual features are object features extracted with a Faster R-CNN \cite{DBLP:journals/pami/RenHG017} -- with the notable exception of \citet{DBLP:journals/corr/abs-1909-02950} who used pooling layers from a CNN. 
A first body of work exploit \textit{single-stream} Transformers in which visual features are incorporated in a BERT-like Transformer: this is the case for VisualBERT \cite{DBLP:journals/corr/abs-1908-03557} and VL-BERT \cite{DBLP:journals/corr/abs-1908-08530}. Other works, such as ViLBERT \cite{DBLP:journals/corr/abs-1908-02265} and LXMERT \cite{DBLP:journals/corr/abs-1908-07490} have investigated \textit{two-stream} approaches: these models employ modality-specific encoders built on standard Transformer blocks, which are then fused into a cross-modal encoder.
Interestingly, none of the aforementioned models have been used for generation tasks such as VQG, tackled in this work. 
\paragraph{Visual Question Generation}
The text-based Question Generation task has been largely studied by the NLP community
\cite{DBLP:conf/emnlp/ZhaoNDK18,DBLP:conf/acl/ScialomPS19}. However, its visual counterpart, Visual Question Generation, has been comparatively less explored than standard well-known multi-modal tasks such as Visual Question Answering
\cite{DBLP:conf/nips/GaoMZHWX15}, Visual Dialog
\cite{DBLP:journals/cviu/DasAZPB17}, or Image Captioning
\cite{DBLP:conf/cvpr/VinyalsTBE15}.

The VQG task was first introduced by \citet{DBLP:journals/corr/YangLFA15} in their Neural Self Talk model: the goal is to gain knowledge about an image by iteratively generating questions (VQG) and answering them (VQA). The authors tackle the task with a simple RNN conditioned on the image, following Image Captioning works such as \citet{DBLP:journals/pami/KarpathyF17}.

Suitable data for the VQG task can come from standard image datasets on which questions have been manually annotated, 
such as $VQG_{COCO}$, $VQG_{Flickr}$, $VQG_{Bing}$ \cite{DBLP:conf/acl/MostafazadehMDM16} , each consisting of 5000 images with 5 questions per image.
Alternatively, VQG samples can be derived from VQA datasets, such as $VQA1.0$ \cite{VQA}, by ``reversing" them (taking images as inputs and questions as outputs).  

A variety of approaches have been proposed. 
\citet{DBLP:conf/acl/MostafazadehMDM16} use a standard Gated Recurrent Neural Network, \emph{i.e.} a CNN encoder followed by a GRU decoder to generate questions. 
\citet{DBLP:conf/ijcai/ZhangQYYZ17} aim at generating, for a given image, multiple visually grounded questions of varying types (\textit{what}, \textit{when}, \textit{where}, etc.); similarly, \citet{DBLP:conf/cvpr/JainZS17} generate diverse questions using Variational Autoencoders.
In \citet{DBLP:conf/cvpr/LiDZCOWZ18}, VQG is jointly tackled along its dual task (VQA), just as \citet{DBLP:journals/corr/YangLFA15}.
In \cite{DBLP:conf/emnlp/PatroKKN18,DBLP:journals/corr/abs-1912-09551}, the image (processed by a CNN) and the caption (processed by a LSTM) are combined in a mixture module, followed by a LSTM decoder to generate the question, leading to state-of-the-art results on the VQG task on $VQA1.0$ data. 
More recently, \citet{patro2020deep} incorporate multiple cues -- place information obtained from PlaceCNN \cite{DBLP:journals/pami/ZhouLKO018}, caption, tags -- and combine them within a deep Bayesian framework where the contribution of each cue is weighted to predict a question, obtaining the current state-of-the-art results on $VQG_{COCO}$.

\section{Model}
\label{section:model}

\begin{figure}\centering\includegraphics[width = .99\columnwidth]{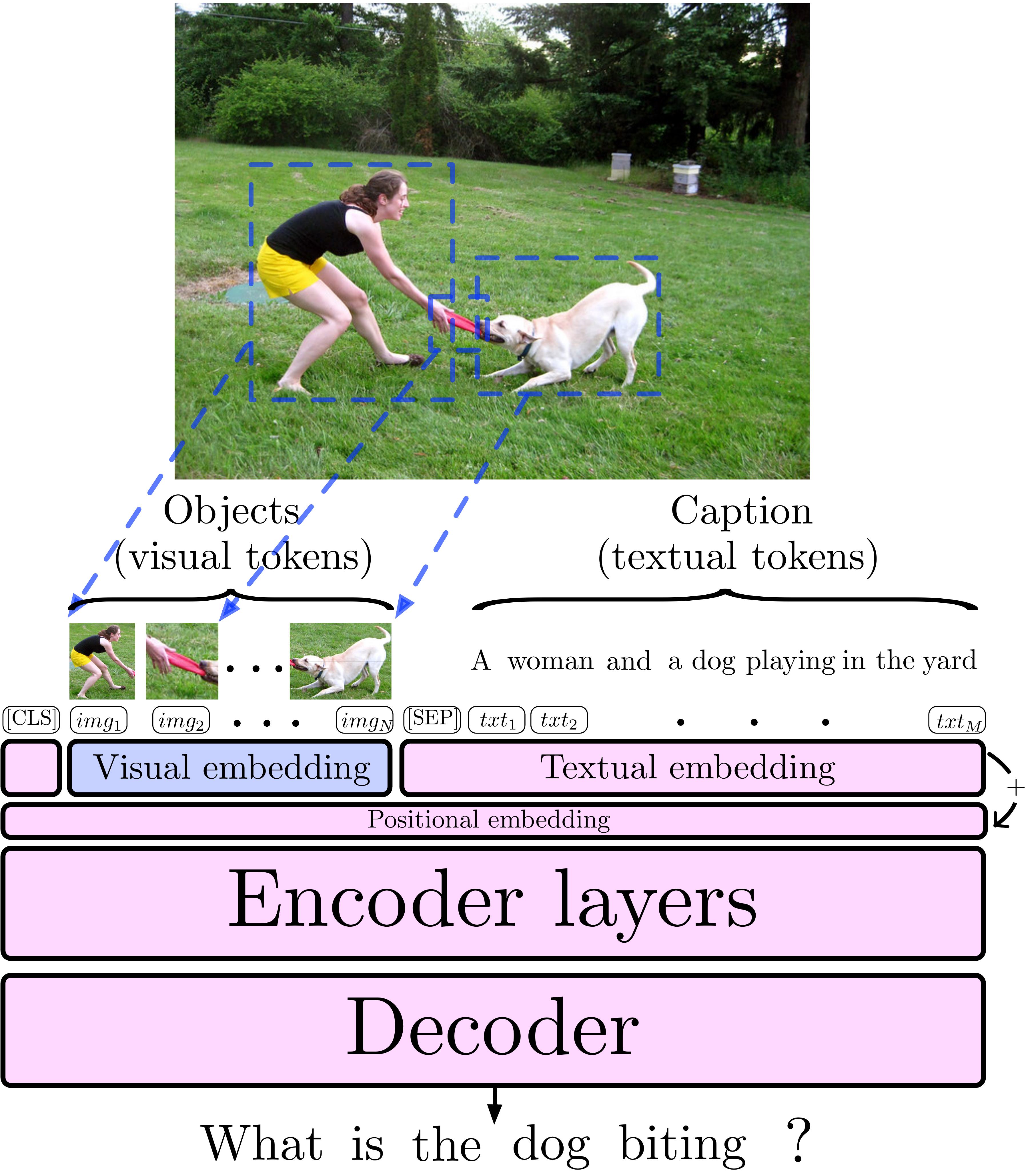}

\caption{Model overview. Captions are encoded via BERT embeddings, while visual embeddings (blue) are obtained via a linear layer, used to project image representations to the embedding layer dimensions.}
\label{model_fig}

\end{figure}

In VQG, the objective is to generate a relevant question from an image and/or its caption. 
The caption $X_{txt}$ is composed of $M$ tokens $txt_1, ..., txt_M$; these tokens can be words or subwords (smaller than word) units depending on the tokenization strategy used. As BERT uses subword tokenization, throughout this paper we will refer to subwords as our tokenization units.

The proposed model is illustrated in Figure~\ref{model_fig}.
In~\ref{rep_im_text}, we detail how images are incorporated in the Transformer framework.
In~\ref{textual_module}, we present \emph{BERT-gen}, a novel approach to use BERT for text generation. 

\subsection{Representing an Image as Text}
\label{rep_im_text}
In this work, we treat textual and visual inputs similarly, by considering both as sequences. Since an image is not a priori sequential, we consider the image $X_{img}$ as a sequence of object regions $img_1, ..., img_N$, as described below.

The images are first processed as in \citet{DBLP:journals/corr/abs-1908-07490}: a Faster-RCNN \cite{DBLP:journals/pami/RenHG017}, pre-trained on Visual Genome \cite{DBLP:journals/ijcv/KrishnaZGJHKCKL17}, detects the $N=36$ most salient regions (those likely to include an object) per image. The weights of the Faster-RCNN are fixed during training, as we use the precomputed representations made publicly available\footnote{\url{https://github.com/peteanderson80/bottom-up-attention}} by \citet{DBLP:conf/cvpr/00010BT0GZ18}.
Each image is thus represented by a sequence of $N=36$ semantic embeddings $f_1, ... f_{N}$ (one for each object region) of dimension $2048$, along with the corresponding bounding box coordinates $b_1, ... b_{N}$ of dimension $4$. With this approach, the BERT attention can be computed at the level of objects or salient image regions; had we represented images with traditional CNN features, the attention would instead correspond to a uniform grid of image regions without particular semantics, as noted in \citet{DBLP:conf/cvpr/00010BT0GZ18}. 
To build an object embedding $o_j$ encoding both the object region semantics and its location in the image, we concatenate $f_j$ and $b_j$ ($j\in [1,N]$). Hence, an image is seen as a sequence of $N=36$ visual representations (each corresponding to an object region) $o_1,..., o_N$. 
Object region representations $o_i$ are ordered by the relevance of the object detected, and the model has access to their relative location in the image through the vectors $b_i$.

To investigate whether our BERT-based model can transfer knowledge beyond language, we consider image features as simple visual tokens that can be presented to the model \emph{analogously} to textual tokens. 
In order to make the $o_j$ vectors (of dimension $2048\hspace{-0.1em}+\hspace{-0.1em}4\hspace{-0.1em}=\hspace{-0.1em}2052$) comparable to BERT embeddings (of dimension $768$), we use a simple linear \textit{cross-modal projection} layer $W$ of dimensions $2052\hspace{-0.1em}\times\hspace{-0.1em}768$. 
The $N$ object regions detected in an image, are thus represented as $X_{img} =  (W.o_1,...,W.o_N)$.
Once mapped into BERT's embedding space with $W$, the image is seen by the rest of the model as a sequence of units with no explicit indication if it is of a textual or visual nature.

\subsection{\emph{BERT-gen}: Text Generation with BERT}
\label{textual_module}
We cast the VQG task as a classic sequence-to-sequence \cite{sutskever2014sequence} framework:

\begin{equation}
    P_{\Theta,W}(Y | X) = \prod_{t=1}^{T}P_{\Theta,W}(y_t|X, y_{<t})
    \label{equation:general_seq2seq}
\end{equation}

\noindent where the input $X=X_{txt}$ in caption-only mode, $X = X_{img}$ in image-only mode, and $X =X_{img} \oplus X_{txt}$ in a multi-modal setup; $Y = {y_1,..., y_T}$ is the question composed of $T$ tokens. $\Theta$ are the parameters of the BERT model;\footnote{We use the smaller architecture released, \texttt{BERT-base} (12 layers), pre-trained on English cased text.} $W$ represents the weights of the linear layer used for projecting visual input to the BERT embedding layer.

As mentioned earlier, BERT is a Transformer \cite{DBLP:conf/nips/VaswaniSPUJGKP17} encoder  pre-trained using the Masked Language Model (MLM) objective: tokens within the text are replaced with a \texttt{[MASK]} special token, and the model is trained to predict them. Since BERT was not trained with an unidirectional objective, its usage for text generation is not straightforward. 

\citet{wang2019bert} demonstrate BERT capacity to infer text, however there method does not allow to train BERT for a text generation task. To that purpose, \citet{liu2019text} propose to stack a Transformer decoder, symmetric to BERT. However, the authors report training difficulties since the stacked decoder is not pre-trained, and propose a specific training regime, with the side-effect of doubling the number of parameters. \citet{dong2019unified} opt for an intermediate step of self-supervised training, introducing an unidirectional loss. 

As detailed below, we propose a relatively simpler, yet effective, method to use BERT \emph{out-of-the-box} for text generation.

\paragraph{Decoder} We simply use the original BERT decoder as is, initially trained to generate the tokens masked during its pre-training phase. It consists in a feed-forward layer, followed by normalization, transposition of the embedding layer, and a softmax over the vocabulary.

\paragraph{Next Token Prediction} At inference time, to generate the first token of the question $y_1$, we concatenate \texttt{[MASK]} to the input tokens $X$, then encode $X \oplus \texttt{[MASK]}$ with the BERT encoder, and feed the output of the encoder to the decoder; $y_1$ is the output of the decoder for the \texttt{[MASK]} token. Subsequently, given $y_1$, we concatenate it to the input tokens and encode $X \oplus y_1 \oplus \texttt{[MASK]}$ to predict the next token $y_2$. This procedure is repeated until the generation of a special token \texttt{[EOS]} signaling the end of the sentence.

\paragraph{Attention Trick} 
As we iteratively concatenate the generated tokens, the BERT \emph{bi-directional} self-attention mechanism would impact, at every new token, the representations of the previous tokens.
To counter that, we use a \emph{left-to-right} attention mask, similar to the one employed in the original Transformer decoder \cite{DBLP:conf/nips/VaswaniSPUJGKP17}.
For the input tokens in $X$, we apply such mask to all the target tokens $Y$ that were concatenated to $X$, so that input tokens can only attend to the other input tokens. Conversely, for target tokens $y_t$, we put an attention mask on all tokens $y_{>t}$, allowing target tokens $y_t$ to attend only to the input tokens and the already generated target tokens. 

This novel method allows to use pre-trained encoders for text generation. In this work, we initialize our model BERT-base parameters. Nonetheless, the methodology can be applied to any pre-trained Transformer encoders such as RoBERTa \cite{liu2019roberta}, or Ernie \cite{sun2019ernie}.

\paragraph{Modality-specific setups}
\label{setups}
The proposed model can be used in either mono- or multi- modal setups. This is accomplished by activating or deactivating specific modules.

\section{Experimental Protocol}
\label{sec:exp_protocol}
Our main objective is to measure whether the textual knowledge encoded in pre-trained BERT can be beneficial in a cross-modal task. 
Thus, we define the three following experimental setups, which we refer to as Step 1, 2, and 3:

\paragraph{1. Caption only} Deactivating the \emph{Visual embedding} module (see Figure~\ref{model_fig}), the model has only access to textual input, \emph{i.e.} the caption. The model is initialized with the BERT weights and trained according to Equation~\ref{equation:general_seq2seq}.

\paragraph{2. Image only} Conversely, deactivating the \emph{Textual embedding} module (see Figure~\ref{model_fig}), the model has only access to the input image, not the caption. To indicate the position $t$ of $img_t$ in the sequence, we sum the BERT positional embedding of $t$ to the visual representation of $img_t$, just as we would do for a text token $txt_t$. The model is initialized with the weights learned during step 1. All \emph{BERT-gen} $\Theta$ weights are frozen, and only the linear layer $W$ is learnable. Hence, \emph{if the model is able to learn to generate contextualized questions w.r.t. the image, it shows that a simple linear layer is enough to bridge the two modalities}.

\paragraph{3. Image + Caption} The full model is given access to both image and caption inputs. In this setup, we separate the two different inputs by a special BERT token \texttt{[SEP]}. Thus, the input sequence for the model takes the form of $\texttt{[CLS]}, img_1,..., img_N, \texttt{[SEP]}, txt_1,..., txt_M$.
In step 1, only \emph{BERT-gen} $\Theta$ parameters are learned, as no image input was given. In step 2, $W$ is trained while keeping $\Theta$ frozen. Finally then, in step 3, we fine-tune the model using both image and text inputs: the model is initialized with the parameters $\Theta$ learned during step 1 and the $W$ learned during step 2, and we unfreeze all parameters.

\paragraph{Ablations} Additionally, we report results obtained with: \emph{Image only (unfreeze)}, where the \emph{BERT-gen} parameters $\Theta$ are not frozen; and \emph{Image+Caption (from scratch)} where the model is learned without the intermediate steps 1 and 2: the \emph{BERT-gen} parameters $\Theta$ are initialized with the weights from pre-trained BERT while $W$ is randomly initialized. 

\subsection{Datasets}

We conduct our experiments using two established datasets for Visual Question Generation. $VQG_{COCO}$ \citet{DBLP:conf/acl/MostafazadehMDM16} contains 2500 training images, 1250 validation images and 1250 test images from MS COCO \cite{DBLP:conf/eccv/LinMBHPRDZ14}; each image has 5 corresponding questions and 5 ground-truth captions.\footnote{Publicly available at \url{https://www.microsoft.com/en-us/download/details.aspx?id=53670}}
The Visual Question Answering ($VQA$) \cite{VQA} dataset can be used to derive VQG data \cite{DBLP:conf/cvpr/LiDZCOWZ18}. The task is reversed: instead of answering the question based on the image (VQA), models are called to generate a relevant question given the image (VQG). Also based on MS COCO, it contains 82783 training images, 40504 validation images and 81434 testing images. In $VQA1.0$,\footnote{Publicly available at \url{https://visualqa.org/vqa_v1_download.html}} 
each image has 3 associated questions.
Since the test set of MS COCO does not contain ground-truth captions, we generated artificial captions for it using NeuralTalk2 \cite{DBLP:journals/pami/KarpathyF17}: for fair comparison, we used exactly the same model\footnote{Publicly available at \url{https://github.com/karpathy/neuraltalk2}} as \citet{DBLP:journals/corr/abs-1912-09551} (MDN-Joint).

\begin{table*}[!ht]
\begin{center}
\begin{tabular}{l|rrrrrrr}
        %   Model 
           & \small{BLEU1} & \small{BLEU2} & \small{BLEU3} & \small{BLEU4} & \small{ROUGE-L} & \small{METEOR} & \small{CIDEr} \\
\hline 
Sample 
%  \cite{DBLP:journals/corr/YangLFA15} 
& 38.8\phantom{0} & -     & -     & -     & 34.2\phantom{0}   & 12.7\phantom{0}  & 13.3\phantom{0} \\
Max 
%  \cite{DBLP:journals/corr/YangLFA15} 
& 59.4\phantom{0} & -     & -     & -     & 49.3\phantom{0}   & 17.8\phantom{0}  & 33.1\phantom{0} \\
MDN-Joint   
%  \cite{DBLP:journals/corr/abs-1912-09551}
& 65.1 & -     & -     & -     & 52.0   & 22.7  & 33.1 \\
\hline 
Cap. only \phantom{a} Step 1             & 75.41 & 56.49 & 43.26 & 32.28 & 66.18   & 26.51  & 43.56 \\
Im. only \phantom{aa} Step 2 (freeze)    & 73.62 & 53.54 & 39.37 & 27.44 & 64.34   & 24.36  & 29.58 \\
Im. only \phantom{aa} Step 2 (unfreeze)  & 73.97 & 55.07 & 42.20 & 31.76 & 65.70   & 26.36  & 41.43 \\
Im. + Cap. \phantom{} Step 3 \phantom{(from scratch)}          & 75.59 & \textbf{56.88} & \textbf{43.96} & \textbf{33.35} & \textbf{66.71}   & \textbf{26.76}  & \textbf{44.99} \\
Im. + Cap.  \phantom{} Step 3 (from scratch) & \textbf{75.84} & 56.42 & 43.53 & 32.85 & 66.30   & 25.92  & 38.81
\end{tabular}
\caption{Quantitative VQG results on $VQA1.0$. We report results from previous works in the upper block, and those obtained by our proposed models in the bottom block.}
\label{table:result-VQA}
\end{center}
\end{table*}

\subsection{Baselines}
\label{seb_seq:Baselines}

We compare the proposed model to the following: 

\paragraph{Sample} \cite{DBLP:journals/corr/YangLFA15} 
Questions are generated by a RNN conditioned on the image: at each generation step, the distribution over the vocabulary is computed and used to sample the next generated word.
This baseline enables to generate diverse questions over the same image, as the word selection process is non-deterministic.

\paragraph{Max} \cite{DBLP:journals/corr/YangLFA15}
Using the above model, selecting words with maximum probability from the computed distribution.

\paragraph{MDN-Joint} \cite{DBLP:journals/corr/abs-1912-09551} State-of-the-art model on $VQA1.0$, based on joint usage of caption and image information.

\paragraph{MC-SBN} \cite{patro2020deep} State-of-the-art on $VQG_{COCO}$. The model jointly leverages on multiple cues (the image, place information, caption, tags) to generate questions.

\subsection{Metrics}
We report the following metrics for all experiments, consistently with previous works: 

\paragraph{BLEU} \cite{papineni2002bleu} A precision-oriented metric, originally proposed to evaluate machine translation. It is based on the counts of overlapping n-grams between the generated sequences and the human references.

\paragraph{ROUGE} \cite{lin-2004-rouge} The recall-oriented counterpart to BLEU metrics, based on n-gram overlaps.

\paragraph{METEOR} \cite{banerjee2005meteor} 
The harmonic mean between precision and recall w.r.t. unigrams. As opposed to the other metrics, it also accounts for stemming and synonymy matching.

\paragraph{CIDEr} \cite{vedantam2015cider} Originally designed for Image Captioning, 
it uses human consensus among the multiple references, favoring rare words and penalizing frequent words. This feature is particularly relevant for our task, as the automatically generated questions often follow similar patterns such as ``What is the [...] ?". Indeed, we verify experimentally (cf Table~\ref{table:result-VQA} and Table~\ref{table:result-VQG-COCO}) that the CIDEr metric is the most discriminant in our quantitative results.

\subsection{Implementation details}
All models are implemented in PyText \cite{aly2018pytext}.
For all our experiments we used a single NVIDIA RTX 2080 Ti GPU, a batch size of $128$ and $5$ epochs. We used the Adam optimizer with the recommended parameters for BERT: learning rate is set at $2e^{-5}$ with a warmup of $0.1$. The most computationally expensive experiment is the step 3 described above: for this model, completion of one epoch demands 30 seconds and 2 minutes for $VQG_{COCO}$ and $VQA$ datasets, respectively. 
Metrics were computed using the Python package released by \citet{DBLP:conf/acl/DuSC17}.\footnote{\url{https://github.com/xinyadu/nqg/tree/master/qgevalcap}}

\section{Results}

\begin{table*}[!ht]
\begin{center}
\begin{tabular}{l|rrrrrrr}

        %   Model 
           & \small{BLEU1} & \small{BLEU2} & \small{BLEU3} & \small{BLEU4} & \small{ROUGE-L} & \small{METEOR} & \small{CIDEr} \\
\hline  
MDN-Joint 
%  \cite{DBLP:journals/corr/abs-1912-09551}
& 36.0 & 24.9 & 16.8 & 10.4 & 41.8   & 23.4  & 50.7 \\
MC-SBN 
%  \cite{patro2020deep} 
& 40.7 & - & - &- &- & 22.6& - \\
% Humans2016 & \textit{86} & -     & -     & -     & -       & \textit{60}  & - \\
\hline 
Cap. only \phantom{a} Step 1    & \textbf{74.58} & 54.94 & 43.33 & 34.36 & 64.09   & \textbf{29.76}  & \textbf{77.70} \\
Im. only \phantom{aa} Step 2 (freeze)    & 69.57 & 49.93 & 38.23 & 29.54 & 61.01   & 27.03  & 57.38 \\
Im. only \phantom{aa} Step 2 (unfreeze)  & 74.34 & \textbf{55.26} & \textbf{43.47} & \textbf{34.41} & \textbf{64.63}   & 29.17  & 72.18 \\
Im. + Cap. \phantom{} Step 3    & 70.96 & 50.83 & 39.20 & 30.29 & 61.87   & 27.65  & 62.77 \\
Im. + Cap.  \phantom{} Step 3 (from scratch) & 64.18 & 42.88 & 30.19 & 20.14 & 56.99   & 23.32  & 30.99 \\
\hline
\emph{Human Performance} & \textit{86} & -     & -     & -     & -       & \textit{60}  & - \\
\end{tabular}
%\vspace{-0.2cm}
\caption{Quantitative VQG results on $VQG_{COCO}$. We report results from previous works in the upper block, and those obtained by the our proposed models in the middle block.
Human Performance is taken from \citet{DBLP:conf/acl/MostafazadehMDM16}.}

\label{table:result-VQG-COCO}
\end{center}
\end{table*}

\begin{table}
\begin{center}
\begin{tabular}{l | lll}
       & Read. & Cap. Rel. & Im. Rel.  \\
\hline 
Caption only     & \textbf{4.9} & \textbf{4.72}* & 4.25* \\
Image only     & 4.77 & 3.87 & 4.32* \\
Image + Caption     &4.89 & 4.06* & \textbf{4.69}*\\
\hline 
\emph{Human} &   \textit{4.83} & \textit{3.64 }& \textit{4.9}        \\
\end{tabular}

\caption{Human evaluation results for three criterions: \textit{readability}, \textit{caption relevance} and \textit{image relevance}. Two-tailed t-test results are reported in comparison to "Human" (*: $p<0.05$).}
\label{human_ev}

\end{center}
\end{table}

In Table~\ref{table:result-VQA}, we report quantitative results for the VQG task on $VQA1.0$. The \textit{Caption only} model already shows strong improvements for all metrics over SOTA models. For this text-only model, the impressive performance can mostly be attributed to BERT, demonstrating once again the benefits of pre-trained language models. 
In our Step 2 (\textit{Image only}), BERT's $\Theta$ parameters are frozen and only those of the cross-modal projection matrix $W$ are learned. Despite using a simple linear layer, the model is found to perform well, generating relevant questions given only visual inputs. 

This suggests that the conceptual representations encoded in pre-trained language models such as BERT can effectively be used beyond text.
Further, we report an additional \textit{Image only} experiment, this time unfreezing the BERT parameters $\Theta$ -- see \emph{Step 2 (unfreeze)} in Table~\ref{table:result-VQA}. As could be expected, since the model is allowed more flexibility, the performance is found to further improve.

Finally, in our third step (\textit{Image + Caption}), we obtain the highest scores, for all metrics. This indicates that the model is able to effectively leverage the combination of textual and visual inputs. Indeed, complementary information from both modalities can be exploited by the self-attention mechanism, making visual and textual tokens interact to generate the output sequences. 
Again, we additionally report the results obtained bypassing the intermediate steps 1 and 2: for the model denoted as \textit{Step 3 (from scratch)} (last row of Table~\ref{table:result-VQA}), $\Theta$ parameters are initialized with the original weights from pre-trained BERT, while the $W$ matrix is randomly initialized. Under this experimental condition, we observe lower performances, a finding that consolidates the importance of the multi-step training procedure we adopted.

In Table~\ref{table:result-VQG-COCO}, we report quantitative VQG results on $VQG_{COCO}$. These are globally consistent with the ones above for $VQA1.0$. However, we observe two main differences. First, a bigger relative improvement over the state-of-the-art. As the efficacy of pre-trained models is boosted in small-data scenarios \cite{radford2018improving}, this difference can be explained by the smaller size of $VQG_{COCO}$. Second, we note that the \emph{Caption only} model globally outperforms all other models, especially on the discriminant CIDEr metric. 
This can be explained by the fact that, in $VQG_{COCO}$, the captions are human-written (whereas they are automatically generated for $VQA1.0$) and, thus, of higher quality; moreover, the smaller size of the dataset could play a role hindering the ability to adapt to the visual modality. 
Nonetheless, the strong performances obtained for \emph{Step 2} compared to the baselines highlight the effectiveness of our method to learn a cross-modal projection even with a relatively small number of training images.

\paragraph{Human Evaluation}

To get more in-depth understanding of our models, we report human assessment results in Table~\ref{human_ev}. We randomly sampled 50 images from the test set of $VQA1.0$. Each image is paired with its caption, the human-written question used as ground-truth, and the output for our three models: \textit{Caption only}, \textit{Image only} and \textit{Image+Caption}.
We asked 3 human annotators to assess the quality of each question using a Likert scale ranging from 1 to 5, for the following criteria: \emph{readability}, measuring how well-written the question is; \emph{caption relevance}, how relevant the question is w.r.t. to the caption; and, \emph{image relevance}, how relevant the question is toward the image. For caption and image relevance, the annotators were presented with only the caption and only the image, respectively. %\textcolor{red}{Add details about the human evaluation if possible (interrater agreement, variance...) }

We observe that all evaluated models produce well-written sentences, as \emph{readability} does not significantly differ compared to human's questions. 
Unsurprisingly, the \emph{Caption only} model shows a higher score for \emph{caption relevance}, while the relatively lower \emph{image relevance} score can be explained by the automatically generated and thus imperfect captions in the $VQA1.0$ dataset. 
Comparatively, the \emph{Image only} model obtains lower \emph{caption relevance} and higher \emph{image relevance} scores; this indicates that the cross modal projection is sufficient to bridge modalities, allowing BERT to generate relevant questions toward the image. Finally, the \textit{Image + Caption} model obtains the best \textit{image relevance} among our models, consistently the quantitative results reported in Tables~\ref{table:result-VQA} and~\ref{table:result-VQG-COCO}. 

\section{Model Discussion}

\begin{figure*}\centering\includegraphics[width =\textwidth]{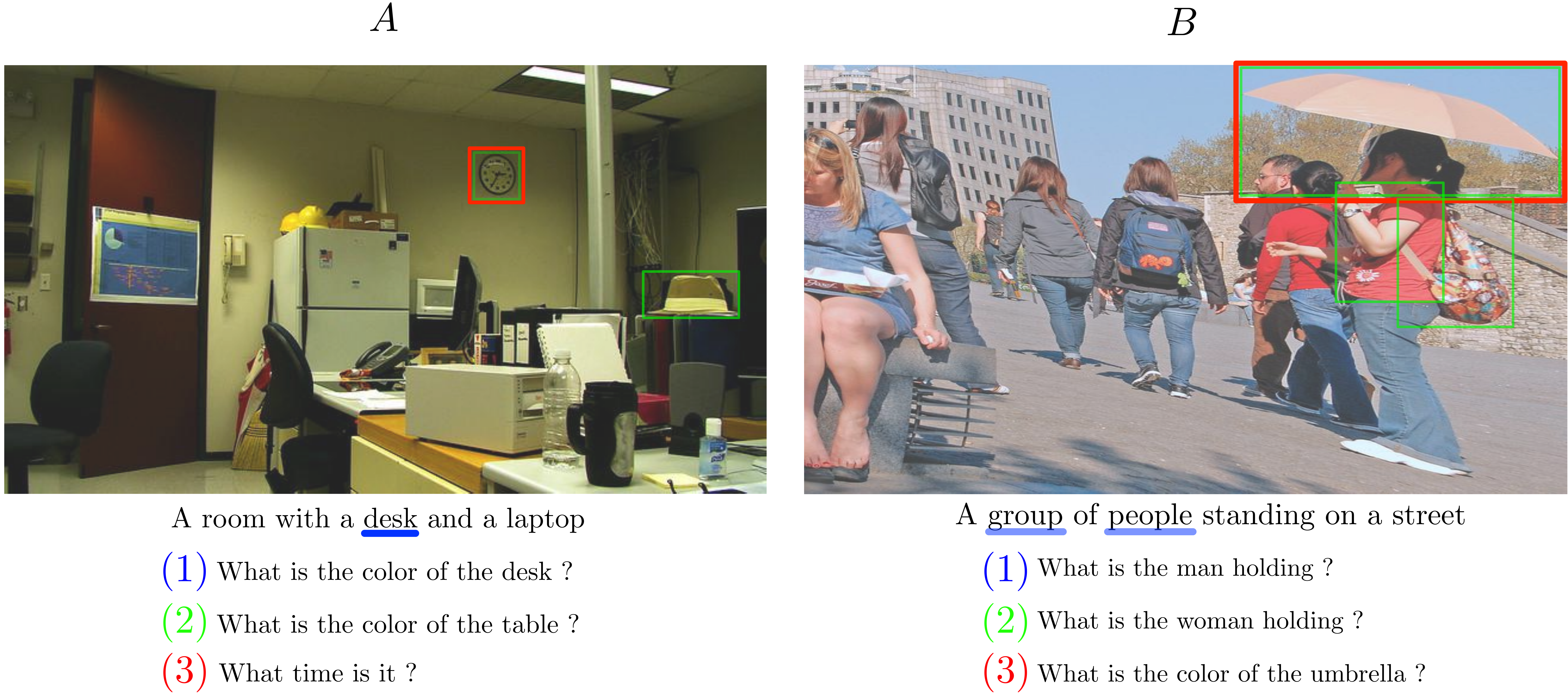}
\caption{Qualitative Analysis. We show the outputs of the three steps of our model, using two samples from the $VQA1.0$ test set. \emph{1)} Caption only; \emph{2)} Image only; \emph{3)} Image + Caption. Words and object regions with maximum attention are underlined and marked, respectively. Color intensity is proportional to attention.}
\label{qualitative}
\end{figure*}

\paragraph{What does the model look at?}
To interpret the behavior of attention-based models, it is useful to look at which tokens are given higher attention \cite{clark2019does}. 
In Figure~\ref{qualitative}, we present two images $A$ and $B$, along with their captions and the three generated questions corresponding to our three experimental setups (\textit{Caption only}, \textit{Image only} and \textit{Image + Caption}). For this analysis, we average the attention vectors of all the heads in the last layer, and highlight the textual and visual tokens most attended by the models.

For both images, the \textit{Caption~only} model attends to salient words in the caption.
The \textit{Image~only} model remains at least as much relevant: on image $A$, it generates a question about a table (with an unclear attention). Interestingly, for image $B$, the \textit{Image~only} model corrects a mistake from step 1: it is a \textit{woman} holding an umbrella rather than a \textit{man}, and the attention is indeed focused on the woman in the image.
Finally, the \textit{Image + Caption} model is able to generate fitting questions about the image, with relatively little relevance to the caption: for image $A$, \textit{Image + Caption} the model generates ``What time is it?" while paying attention to the clock; for image $B$, \textit{Image + Caption} generates ``What is the color of the umbrella ?", focusing the attention on the umbrella. The captions of either samples include no mentions of clocks or umbrellas, further indicating effective alignment between visual and textual representations.

\begin{figure}\centering\includegraphics[width = \columnwidth]{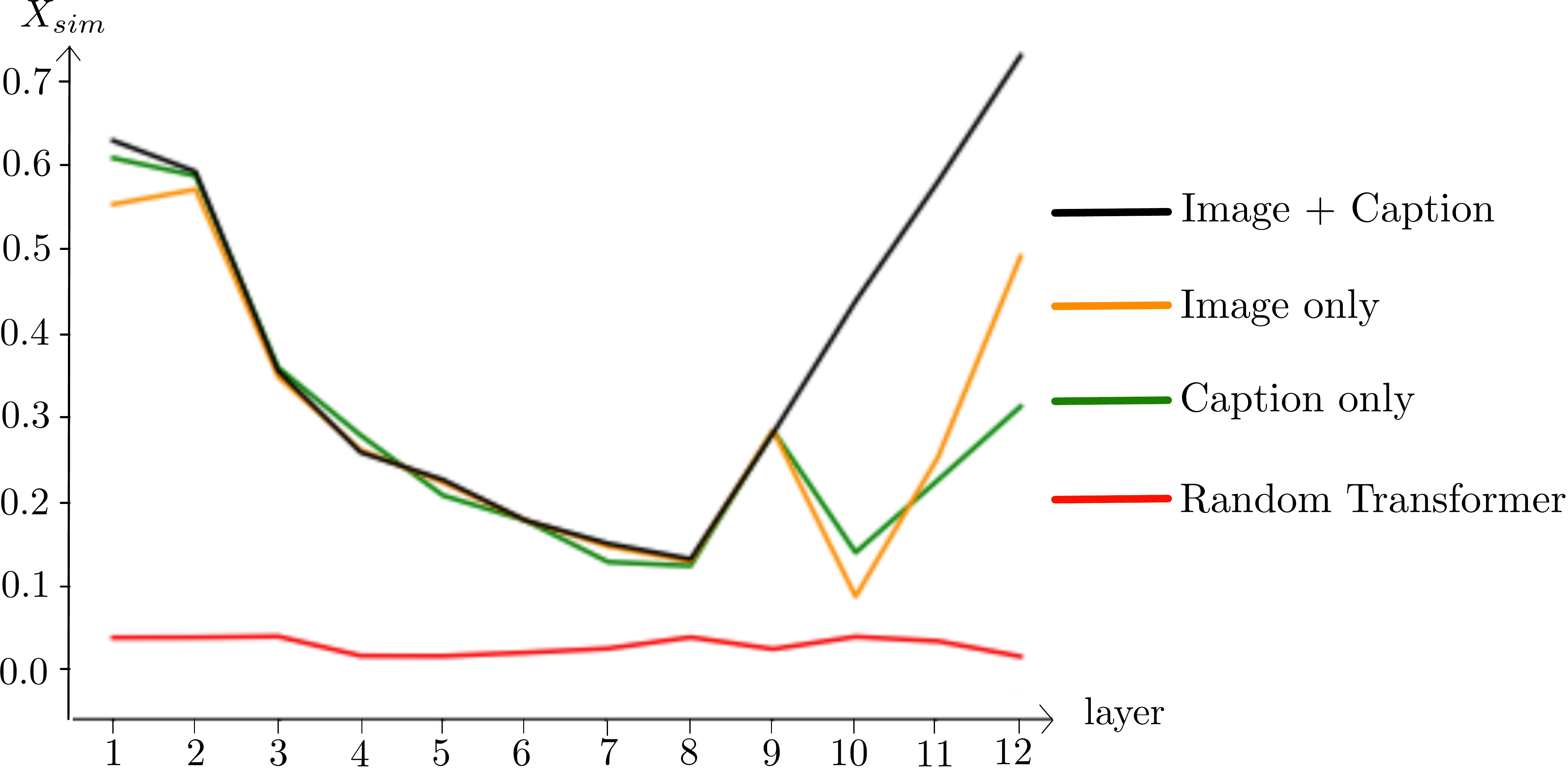}
\caption{Cross-modal similarity $X_{sim}$ between images in $VQG_{COCO}$ and corresponding captions at each BERT encoding layer. Captions and images are embedded here using the \texttt{[CLS]} special token. 
}
\label{quantitative}
\end{figure}

\paragraph{Cross-modal alignment}
We carry out an additional experiment to analyze the cross-modal alignment for each version of our model.

Figure~\ref{quantitative} shows the \textit{cross-modal} similarity $X_{sim}$ for different model scenarios,  computed at each BERT-base layer from 1 to 12. 
We define the cross-modal similarity $X_{sim}$ as the cosine similarity between the vector representations of both modalities. 
These vectors are the two continuous space representations from a model when given as input either \emph{i)} an image, or \emph{ii)} its corresponding caption.
We represent these captions and images vectors with the special BERT token \texttt{[CLS]}, following previous works \cite{DBLP:conf/nips/ReifYWVCPK19} where \texttt{[CLS]} is used to represent the entire sequence. 

Reported values are averaged over all examples of $VQG_{COCO}$ test set.
In addition to the setups described in Section~\ref{sec:exp_protocol} (\emph{Caption-only}, \emph{Image-only} and \emph{Image + Caption}), we also report $X_{sim}$ for \textit{Random Transformer}, a BERT architecture with random weights --- as expected, its $X_{sim}$ is close to zero. $W$ is set at random for models where visual data has not been used (\textit{Random Transformer}, \textit{Caption only}).

All the other models are based on BERT. As suggested by \citet{tenney2019bert}, the first layers in BERT tend to encode lower-level language information. This might explain why the models show similar $X_{sim}$ scores up to the 9th layer, and diverge afterwards: the weights for those layers remain very similar between our fine-tuned models.

For the last layer ($l=12$), we observe that $\textit{Caption only} < \textit{Image only} < \textit{Image + Caption}$. The \textit{Caption only} model has never seen images during training, and therefore is not able to encode semantic information given only images as input. Still, its reported $X_{sim} > 0$ can be attributed to the fact that, when fine-tuned on VQG during Step 1, \emph{BERT-gen} encodes task-specific information in the \texttt{[CLS]} token embedding (\emph{e.g.} a question ends with a ``?" and often begins with ``What/Where/Who"). 
$\textit{Image only} > \textit{Caption only}$ comes from the learning of the cross-modal matrix $W$. However, since BERT is not fine-tuned, the model learns a \textit{contortion} allowing it to align text and vision.
Finally, \textit{Image + Caption} $>$ \textit{Image only} comes from BERT's fine-tuning, contributing to an increase in the observed gap, and its emergence in earlier layers.

\section{Conclusion and Perspectives}
We investigated whether the abstractions encoded in a pre-trained BERT model can generalize beyond text.
We proposed \emph{BERT-gen}, a novel methodology that allows to directly generate text from \emph{out-of-the-box} pre-trained encoders, either in mono- or multi- modal setups. 
Moreover, we applied \emph{BERT-gen} to Visual Question Generation, obtaining state-of-the-art results on two established datasets. We showed how a simple linear projection is sufficient to effectively align visual and textual representations.

\bibliography{inlg2020}
\bibliographystyle{acl_natbib}

\end{document}